\PassOptionsToPackage{unicode}{hyperref}
\PassOptionsToPackage{hyphens}{url}
\documentclass[
]{article}
\usepackage{xcolor}
\usepackage{amsmath,amssymb}
\usepackage{iftex}
\ifPDFTeX
  \usepackage[T1]{fontenc}
  \usepackage[utf8]{inputenc}
  \usepackage{textcomp} 
\else 
  \usepackage{unicode-math} 
  \defaultfontfeatures{Scale=MatchLowercase}
  \defaultfontfeatures[\rmfamily]{Ligatures=TeX,Scale=1}
\fi
\usepackage{lmodern}
\ifPDFTeX\else
\fi
\IfFileExists{upquote.sty}{\usepackage{upquote}}{}
\IfFileExists{microtype.sty}{
  \usepackage[]{microtype}
  \UseMicrotypeSet[protrusion]{basicmath} 
}{}
\makeatletter
\@ifundefined{KOMAClassName}{
  \IfFileExists{parskip.sty}{%
    \usepackage{parskip}
  }{
    \setlength{\parindent}{0pt}
    \setlength{\parskip}{6pt plus 2pt minus 1pt}}
}{
  \KOMAoptions{parskip=half}}
\makeatother
\usepackage{longtable,booktabs,array}
\usepackage{calc} 
\usepackage{etoolbox}
\makeatletter
\patchcmd\longtable{\par}{\if@noskipsec\mbox{}\fi\par}{}{}
\makeatother
\IfFileExists{footnotehyper.sty}{\usepackage{footnotehyper}}{\usepackage{footnote}}
\makesavenoteenv{longtable}
\usepackage{graphicx}
\makeatletter
\newsavebox\pandoc@box
\newcommand*\pandocbounded[1]{
  \sbox\pandoc@box{#1}%
  \Gscale@div\@tempa{\textheight}{\dimexpr\ht\pandoc@box+\dp\pandoc@box\relax}%
  \Gscale@div\@tempb{\linewidth}{\wd\pandoc@box}%
  \ifdim\@tempb\p@<\@tempa\p@\let\@tempa\@tempb\fi
  \ifdim\@tempa\p@<\p@\scalebox{\@tempa}{\usebox\pandoc@box}%
  \else\usebox{\pandoc@box}%
  \fi%
}
\def\fps@figure{htbp}
\makeatother
\providecommand{\tightlist}{%
  \setlength{\itemsep}{0pt}\setlength{\parskip}{0pt}}
\usepackage[]{natbib}
\usepackage[letterpaper,margin=1in]{geometry}
\usepackage{unicode-math}
\usepackage{longtable,booktabs,array,calc}
\usepackage{newunicodechar}
\newunicodechar{–}{\textendash}
\newunicodechar{—}{\textemdash}
\newunicodechar{≈}{\ensuremath{\approx}}
\newunicodechar{⇒}{\ensuremath{\Rightarrow}}
\newunicodechar{→}{\ensuremath{\rightarrow}}
\newunicodechar{×}{\ensuremath{\times}}
\newunicodechar{≤}{\ensuremath{\leq}}
\newunicodechar{≥}{\ensuremath{\geq}}
\newunicodechar{∼}{\ensuremath{\sim}}
\newunicodechar{≪}{\ensuremath{\ll}}
\newunicodechar{≫}{\ensuremath{\gg}}
\newunicodechar{∈}{\ensuremath{\in}}
\newunicodechar{≲}{\ensuremath{\lesssim}}
\newunicodechar{σ}{\ensuremath{\sigma}}
\newunicodechar{∂}{\ensuremath{\partial}}
\newunicodechar{√}{\ensuremath{\surd}}
\usepackage{bookmark}
\IfFileExists{xurl.sty}{\usepackage{xurl}}{} 
\makeatletter
\@ifundefined{xmpquote}{}{}
\makeatother
\hypersetup{
  pdftitle={When Do Conservation Laws Survive Learned Representations? Certified Horizons for Latent World Models},
  pdfauthor={Hongbo Wang},
  hidelinks,
  pdfcreator={LaTeX via pandoc}}

\title{When Do Conservation Laws Survive Learned Representations?
Certified Horizons for Latent World Models}
\author{Hongbo Wang \\
  \small Department of Mathematics, Stony Brook University, Stony Brook, NY 11794, USA}
\date{}

\begin{document}
\maketitle
\begin{abstract}
We ask a representation-learning question about physical world models:
when does a conservation law remain \emph{certifiable} after a model
learns a latent representation? A certified horizon bounds --- in
advance, from measurable model defects --- how many steps a rollout
provably stays on a physical invariant's level set. The key design
choice is \emph{what} is certified: not a learned latent Hamiltonian or
a learned scalar witness (a model can conserve either while drifting in
true energy), but the \textbf{decoded physical invariant} obtained by
decoding the latent state and evaluating the known invariant. Around
this object we derive shell-horizon certificates whose budget decomposes
into representation, readout, and latent-dynamics defects, with a
monotone \emph{alignment bridge} through which a soft learned witness
yields a certified horizon for the decoded invariant, and test them
across state, learned-lift, and pixel observations on conservative
systems. Conservation certificates \emph{can} survive learned
representation, but not all geometric priors survive equally. Hard
canonical symplectic structure yields the longest horizons in known
phase coordinates yet does not cross a learned chart, whereas a
controlled-Lipschitz-aligned soft invariant survives in the nonlinear
learned-representation settings we test --- two lift systems, with the
gain growing with nonlinearity, and pixels. Pixel certification is
recovered on a readout-stable sub-tube, and the Kepler problem exposes a
geometric boundary. The central object is therefore not a latent
Hamiltonian, but a decoded physical invariant whose robustness to
representation learning can be measured, certified, and falsified.
\end{abstract}

\subsection{1. Introduction}\label{introduction}

A world model is only deployable if it can say \emph{when to stop
trusting itself}, and modern world models earn their leverage by not
predicting in observation space at all: they learn a latent
representation and roll forward there. That is exactly what makes the
trust question subtle. For physical systems there is a natural currency
for trust --- conserved quantities: if a rollout drifts off the energy
shell after \(n\) steps, then \(n\) is a concrete, checkable horizon.
But conservation is a statement about \emph{physical} coordinates, and a
latent world model has, by construction, replaced those coordinates with
learned ones. The question of this paper is therefore a
representation-learning question: \emph{when does a conservation law
remain} \textbf{certifiable} \emph{after a model learns a latent
representation} --- bounded ahead of time from measurable model defects,
rather than merely observed after the fact?

Two facts make the question non-trivial. First, the dominant guarantees
for learned dynamics are \emph{pointwise and spectral} --- they bound
one-step error growth through a Jacobian or Koopman spectrum --- so they
decay geometrically and certify how fast trajectories \emph{separate},
not whether a conserved \emph{structure} is maintained over many steps.
Conservation offers a different, slower currency: a spectral horizon
bounds how fast trajectories diverge (decaying like
\(\log(1/\epsilon)/\lambda\)), whereas a conservation horizon bounds how
long a conserved scalar stays put --- often far longer. Second, and this
is the crux, the moment the model is latent one must decide \emph{what}
is being conserved. The learned latent Hamiltonian \(H_\theta\) and a
learned scalar witness \(C_\omega\) are convenient and architecturally
natural, but neither is the physical invariant: a model can conserve its
own internal scalar perfectly while the true energy drifts.
Representation learning does not merely complicate certification --- it
creates an \emph{object problem} about what the certificate is attached
to in the first place. Latent world models may preserve
\emph{something}, but unless that something aligns with a decoded
physical invariant, it is not a certificate.

Our answer fixes the certified object to the \textbf{decoded physical
invariant} \(H^\star(\Pi D_\psi z)\): decode the latent state
(\(D_\psi\)), read out physical coordinates (\(\Pi\)), and evaluate the
\emph{known} physical invariant \(H^\star\). Concretely, for a pendulum
observed as pixels, \(H^\star\) is the true mechanical energy: the model
never receives it during training or rollout, and we evaluate it only
after decoding the latent state and reading out \((q,p)\) --- so the
certificate measures whether \emph{physical} energy, not a
model-internal surrogate, stays on its shell. Figure 1 makes this the
spine of the paper --- \(H_\theta\) and \(C_\omega\) are
witnesses/proxies on a side branch with no path to the certified object;
certificates are evaluated only where the state pipeline and the
latent-decoded pipeline converge on \(H^\star(\Pi D_\psi z)\). Two
principles then govern every claim below: the certified object is
\emph{always} the decoded physical invariant, and every success or
failure is attributed honestly along a state \(\to\) lift \(\to\) pixel
ladder of increasing representational distance from physical
coordinates.

\begin{figure}
\centering
\includegraphics[width=0.7\linewidth,height=\textheight,keepaspectratio,alt={Certificates are evaluated on the decoded physical invariant H\^{}\textbackslash star(\textbackslash Pi D\_\textbackslash psi z), where the state pipeline and the latent-decoded pipeline converge --- not on the learned latent Hamiltonian H\_\textbackslash theta or the witness C\_\textbackslash omega, drawn as a demoted side branch with no path to the certified object.}]{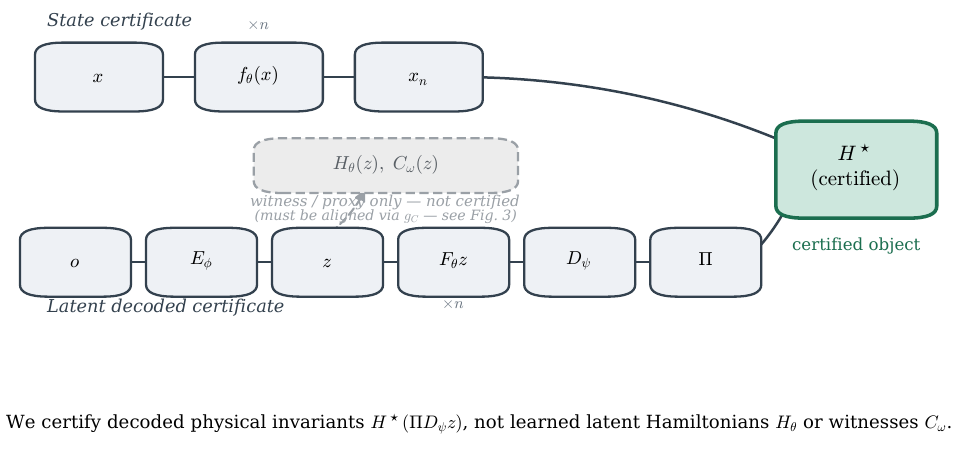}
\caption{Certificates are evaluated on the decoded physical invariant
\(H^\star(\Pi D_\psi z)\), where the state pipeline and the
latent-decoded pipeline converge --- not on the learned latent
Hamiltonian \(H_\theta\) or the witness \(C_\omega\), drawn as a demoted
side branch with no path to the certified object.}
\end{figure}

With the object fixed, the scientific content becomes a question about
\emph{representation robustness}: which structural priors let a
conservation certificate survive the move from physical state to a
learned chart? We state the thesis once: \emph{conservation-aligned
certificates can survive learned representation, but not all geometric
priors survive equally.} Hard canonical symplectic structure --- the
default inductive bias of Hamiltonian world models --- works in known
phase coordinates but does not automatically cross a learned chart,
because symplecticity constrains the \emph{form} of the flow (it
preserves a symplectic form), not \emph{which} scalar equals physical
energy. A \emph{soft} learned invariant, trained to be conserved in the
latent space and then \emph{aligned} to physical energy through a
controlled-Lipschitz bridge, is more robust to exactly the
representation change that defeats the hard prior. Pixel observation
adds an orthogonal \emph{perceptual} bottleneck --- a decoder/readout
sits between latent and physical coordinates --- and a singular system
(the Kepler two-body problem) exposes a purely geometric limit to
joint-invariant certification. The practical reading is conditional, not
dismissive: certify in coordinates where the latent symplectic form is
the physical one, or pair the hard structure with an explicit alignment
of the conserved scalar --- otherwise the prior protects a \emph{shadow}
quantity that has drifted from \(H^\star\) once the chart is learned.

We test this across three representations on conservative systems
(pendulum, harmonic oscillator, Kepler two-body), holding the certified
object fixed and varying only what the model sees. Four findings,
attributed in full in §5: (i) in known canonical state, hard symplectic
structure is the clear winner --- on Kepler the joint-invariant shell is
non-vacuous in \(5/5\) seeds for symplectic versus \(0/5\) for plain, a
\(\sim 6.5\times\) smaller per-step invariant drift; (ii) under a
learned lift the hard prior collapses to the unconstrained baseline ---
separable Verlet and non-separable SympNet alike --- while the soft
invariant survives on both nonlinear lift systems we test
(structure-gain \(\approx 1.3\) on the pendulum, \(\approx 2.3\) on the
Duffing double well; \(P(\text{soft}>\text{plain})=0.88\) over eight
seeds on each), with a positive witness-alignment audit on the pendulum
(mean \(R^2\approx 0.98\), controls \(\approx 0\)); (iii) under pixels
the stabilized soft witness recovers a non-vacuous decoded-energy
certificate on a \emph{readout-stable sub-tube} --- beating plain in
\(8/10\) seeds at \(\epsilon=2.0\), non-vacuous in \(9/10\) (plain
\(4/10\)), alignment-positive in \(10/10\), excluded fraction
\(\le 2.3\%\); and (iv) the Kepler \emph{lift} hits an intrinsic
near-periapsis stiffness floor that survives capacity scaling and every
legal metric re-charting we tried --- a clean negative-with-mechanism,
not a structure failure.

\textbf{Contributions.} 1. \textbf{A representation-robustness framing
for conservation certificates.} We fix the certified object to the
decoded physical invariant \(H^\star(\Pi D_\psi z)\) --- not
\(H_\theta\) or \(C_\omega\) --- and study when a certificate built on
it survives a learned chart (Figure 1; §3--4). 2. \textbf{Certificate
theory} (§3): a unified family of decoded-invariant horizon bounds --- a
state shell, a latent decoded-energy shell with an explicit three-way
error budget, a controlled-Lipschitz \emph{alignment bridge} that
bridges a soft witness to the certified decoded invariant, and a vector
joint-invariant shell. 3. \textbf{A hard-vs-soft survival result under
learned representation} (§5): hard symplectic wins in canonical
coordinates (\(5/5\) vs \(0/5\)) but collapses across a learned lift,
while a controlled-Lipschitz-aligned soft invariant survives on both
lift systems tested (pendulum \(\approx 1.3\), Duffing \(\approx 2.3\);
\(P=0.88\) each); \emph{monotonicity alone is insufficient}. 4.
\textbf{Two honest representation-limit results}: a pixel certificate
recovered only on a readout-stable sub-tube (excluded \(\le 2.3\%\)),
and a Kepler-lift geometric boundary robust to capacity and legal
re-charting.

\begin{center}\rule{0.5\linewidth}{0.5pt}\end{center}

\subsection{2. Related work}\label{related-work}

These four lines each supply useful structure for \emph{learning}
dynamics; our contribution is orthogonal --- we ask which physical
invariant is actually \emph{certified} after encoding and decoding, and
how hard versus soft conservation priors behave under a learned
representation. (We do not claim prior lines ignore conservation, that
symplectic models are useless, or that we are first to use invariants.)

\textbf{Structure-preserving and invariant learning.} Two related lines
build conservation into learned dynamics. Hamiltonian/symplectic models
impose phase-space structure: Hamiltonian Neural Networks
\citep{greydanus2019hamiltonian}, Hamiltonian Generative Networks
\citep{toth2020hamiltonian} (which infer a latent phase space from
images), SympNets \citep{jin2020sympnets}, and the
geometric-numerical-integration / backward-error tradition
\citep{hairer2006geometric, reich1999backward} that explains symplectic
integrators' long-time energy behavior. A second line learns invariants
directly: meta-learned conserved quantities \citep{alet2021noether},
data-driven conservation-law discovery \citep{kaiser2018discovering},
and Koopman-operator methods whose unit-eigenvalue eigenfunctions
express conserved observables \citep{lusch2018deep} --- our soft
``B-self'' witness \(C_\omega\) belongs here. \emph{Distinction.} These
methods learn or impose structure to improve the \emph{model}; they do
not ask what is \emph{certified} once the model operates on a learned
representation. Our certified object is the decoded physical invariant
\(H^\star(\Pi D_\psi z)\), not \(H_\theta\) or a learned witness; and a
learned invariant is a \textbf{witness, not a certificate} unless
aligned to the decoded invariant through a bi-Lipschitz bridge (§3.3),
where monotonicity alone is insufficient. (HGN is the closest structural
template; we treat ``learning a latent phase space from pixels'' as
established prior art.)

\textbf{Equivariant / geometric deep learning.} A second line imposes
symmetry through equivariant networks
\citep{satorras2021en, cohen2016group}. \emph{Distinction.} Our focus is
not generic equivariance but conservation-aligned \emph{certificate
objects} under learned latent representations; equivariance constrains
how a representation transforms, whereas we ask which decoded physical
invariant admits a horizon certificate after encoding/decoding. (A
separate spacetime-equivariant world-model direction is related but not
central here; this line is positioning only.)

\textbf{Certified / long-horizon world models.} A third line bounds how
far a learned model can be trusted: spectral and Lyapunov-style
certificates bound one-step error growth; conformal and reachability
methods calibrate prediction sets or safe tubes
\citep{angelopoulos2021gentle}; and a recent line establishes spectral
certified horizons \(T_j(\epsilon)\sim\log(1/\epsilon)/\lambda_j\) for
equivariant world models \citep{wang2026certified}, subsequently made
distribution-free via conformal orbit-valid trust horizons
\citep{wang2026conformal}. \emph{Distinction.} Existing certificates are
largely \emph{pointwise/spectral} and decay geometrically; this paper
builds \textbf{conservation-aligned} horizons tied to a decoded physical
invariant, whose budget is a per-step invariant drift, not a Jacobian
spectrum (the spectral line is the complementary
\(T_{\rm state}^{\rm spec}\) baseline).

\begin{quote}
\textbf{Positioning.} The novelty is not ``using Hamiltonian priors'';
it is \textbf{certifying decoded physical invariants through learned
representations}, and showing that hard canonical symplectic structure
is coordinate-sensitive (it wins in known phase coordinates but does not
automatically survive a learned chart) while a
controlled-Lipschitz-aligned soft invariant does.
\end{quote}

\begin{center}\rule{0.5\linewidth}{0.5pt}\end{center}

\subsection{3. Conservation-aligned
certificates}\label{conservation-aligned-certificates}

\subsubsection{3.1 Setup}\label{setup}

Let \(H^\star\) be a known physical invariant of a continuous system,
\(\Sigma_c = \{x : H^\star(x) = c\}\) an energy shell, and
\(K_{c,\rho}\) a tube around it. The certificate is the largest number
of model steps for which a rollout provably stays within tolerance
\(\epsilon\) of the shell. Every bound below has the same shape --- a
tolerance budget \(m_{c,\rho}\,\epsilon\) minus an initialization
defect, divided by a per-step drift --- and differs only in \emph{which}
defects enter as the representation becomes more abstract. We write
\(m_{c,\rho} = \inf_{x\in K_{c,\rho}}\lVert\nabla H^\star(x)\rVert > 0\)
for the shell's sensitivity and use \(\lfloor\cdot\rfloor\) since
horizons are integer step counts.

\subsubsection{3.2 State and latent decoded-energy shells (Propositions
A, B)}\label{state-and-latent-decoded-energy-shells-propositions-a-b}

\textbf{Proposition A (state shell).} With one-step state defect
\(\delta_{\rm state} = \sup_{K}\lvert H^\star(f_\theta x) - H^\star(x)\rvert\)
and initialization defect \(\epsilon_0^{\rm state}\), telescoping and
the mean-value theorem give
\(\lvert H^\star(x_n) - H^\star(x_0)\rvert \le n\,\delta_{\rm state}\)
and
\[T_{\rm shell}^{\rm state} = \left\lfloor \frac{m_{c,\rho}\,\epsilon - \epsilon_0^{\rm state}}{\delta_{\rm state}} \right\rfloor .\]

\textbf{Proposition B (latent decoded-energy shell).} For
\(\hat x_n = \Pi D_\psi F_\theta^{\,n} E_\phi(o_0)\), with tube
decoded-energy drift
\(\delta_{\rm phys}^{\rm tube} = \sup_{z\in K_z}\lvert H^\star(\Pi D_\psi F_\theta z) - H^\star(\Pi D_\psi z)\rvert\)
and initial decoded defect
\(\epsilon_0 = \lvert H^\star(\Pi D_\psi E_\phi o_0) - H^\star(x_0)\rvert\),
\[\lvert H^\star(\hat x_n) - H^\star(x_0)\rvert \le \epsilon_0 + \epsilon_{\Pi,\rm num} + n\,\delta_{\rm phys}^{\rm tube}, \qquad T_{\rm shell}^{\rm latent} = \left\lfloor \frac{m_{c,\rho}\,\epsilon - \epsilon_0 - \epsilon_{\Pi,\rm num}}{\delta_{\rm phys}^{\rm tube}} \right\rfloor .\]
The budget splits three ways: an \textbf{initial embedding error}
\(\epsilon_0\) (representation), a \textbf{decoder/readout error}
\(\epsilon_{\Pi,\rm num}\) (readout), and a \textbf{latent-dynamics
drift} \(\delta_{\rm phys}^{\rm tube}\) (dynamics). In the lift setting
\(\epsilon_{\Pi,\rm num}\approx 0\) and the certificate is
dynamics-limited; in pixels the decoder/readout term can dominate
(§5.3); in the Kepler lift the embedding term \(\epsilon_0\) dominates
(§5.4). Full proofs are in Appendix A.

\subsubsection{3.3 The alignment bridge (Proposition
C)}\label{the-alignment-bridge-proposition-c}

A soft witness \(C_\omega\) is not a priori the physical energy; it must
be \emph{aligned}. Let
\(\delta_C = \sup\lvert C_\omega(Fz) - C_\omega(z)\rvert\), let \(g_C\)
be an \(L_g\)-Lipschitz calibration, and define the decoder-side defect
\(\eta_D = \sup_z\lvert \Psi(z) - g_C(C_\omega z)\rvert\) where
\(\Psi(z) = H^\star(\Pi D_\psi z)\). The decoder-side bound (C1) reads
\[\lvert \Psi(F^n z) - \Psi(z)\rvert \le 2\eta_D + L_g\, n\,\delta_C, \qquad T_{\rm align} = \left\lfloor \frac{m_{c,\rho}\,\epsilon - 2\eta_D}{L_g\,\delta_C} \right\rfloor ,\]
with a physical-initial variant (C2) absorbing an encoder-side defect.
The load-bearing condition is \textbf{bi-Lipschitz} calibration,
\(0 < \kappa \le g_C' \le L_g\): only then is \(C_\omega\)
quantitatively comparable to the decoded-energy level sets. This
upgrades the soft witness from a regularizer to a theorem-shaped
\emph{bridge} to the decoded invariant, and it is why affine is too
rigid and \emph{uncontrolled} monotonicity is too loose. We stress:
\(g_C\circ C_\omega\) is the \emph{bridge} (a witness alignment); the
certified object remains \(\Psi\).

\subsubsection{3.4 The joint-invariant
shell}\label{the-joint-invariant-shell}

For systems with more than one conserved quantity we certify the
\emph{vector} invariant. For Kepler we use the normalized
\(\tilde I = (\tilde H, \tilde L)\) (energy and angular momentum, each
standardized). With per-step drift
\(\delta_I = \sup\lVert \tilde I(Fz) - \tilde I(z)\rVert\) and
worst-case sensitivity
\(\sigma_* = \inf_{x\in K}\sigma_{\min}(D\tilde I(x))\) --- guarded by a
rank condition \(\operatorname{rank} D\tilde I = 2\) on the tube ---
\[T_{\rm joint}(\epsilon) = \left\lfloor \frac{\sigma_*\,\epsilon - \epsilon_0}{\delta_I} \right\rfloor .\]
The rank/\(\sigma_*\) guard later makes a metric re-charting
\emph{legal}: a chart cannot buy horizon by collapsing \(\sigma_*\)
(§5.4).

\subsubsection{3.5 Conditional outlook
(D--F)}\label{conditional-outlook-df}

Backward-error / shadow-Hamiltonian alignment (D), near-integrable
action drift (E), and shadowing (F) are stated as program-level /
conditional theorems (Appendix A.5), not load-bearing for the empirical
claims here. Proposition D supplies the discussion's intuition:
\emph{symplecticity constrains the geometry of the flow, not which
scalar equals physical energy}, so a symplectic learned map need not
preserve \(H^\star\) under an arbitrary chart.

\begin{center}\rule{0.5\linewidth}{0.5pt}\end{center}

\subsection{4. Methods}\label{methods}

This section says, concretely, what is trained and how a certificate is
computed from it. Nothing here uses the ground-truth invariant as a
training signal for the soft witness; \(H^\star\) enters only at
\emph{evaluation}.

\subsubsection{4.1 Systems and data}\label{systems-and-data}

We instantiate the framework on three conservative systems: the simple
\textbf{pendulum}, the \textbf{harmonic oscillator}, and the
\textbf{Kepler two-body} problem --- the last being the only
\emph{singular} system (an inverse-square potential whose invariant
stiffens near periapsis), which is what later exposes a geometric
boundary (§5.4). For each system the physical invariant \(H^\star\)
(and, for Kepler, the angular momentum \(L\)) is known in closed form
and is used only for evaluation. Trajectories are integrated from
initial conditions sampled across a band of energy shells. To keep the
certifier honest we use disjoint seed triples per run --- a training
seed, an evaluation seed, and a separate calibration seed for the
alignment audit --- so the data a certificate is read on is never the
data it was fit on. Runs use \(8\) seeds (alignment bridge), \(10\)
seeds (pixel recovery), and \(3\) seeds (Kepler state/lift), with
deterministic kernels and pinned threads. These systems are deliberately
small: a known invariant and a clean readout are what let us isolate the
certificate \emph{object} and the representation gap exactly. Scaling to
realistic systems without known readouts is not the claim of this paper
--- exact invariants and readouts are precisely what make a
certified-object claim testable in the first place.

\subsubsection{4.2 World-model pipeline}\label{world-model-pipeline}

We use the standard latent-rollout pipeline. An observation \(o\) is
encoded to a latent \(z = E_\phi(o)\), advanced by a learned latent map
\(F_\theta\), decoded by \(D_\psi\), and read out to physical
coordinates by \(\Pi\); an \(n\)-step rollout decodes to an observation
\(\hat o_n = D_\psi F_\theta^{\,n} E_\phi(o_0)\) and reads out to the
physical state
\[\hat x_n \;=\; \Pi\,\hat o_n \;=\; \Pi\, D_\psi\, F_\theta^{\,n}\, E_\phi(o_0).\]
The three representations differ only in what plays the role of \(o\)
and how much of the pipeline is learned. In \textbf{state} the model is
handed canonical coordinates \((q,p)\) and \(\Pi\) is the identity, so
the readout term vanishes (\(\epsilon_{\Pi,\rm num}=0\)). In
\textbf{lift} an encoder learns a latent chart from low-dimensional
observations and a clean readout recovers \((q,p)\), so
\(\epsilon_{\Pi,\rm num}\approx0\) and the certificate is
dynamics-limited. In \textbf{pixel} the observation is a
high-dimensional image and \(D_\psi,\Pi\) form a learned decoder/readout
that can be the dominant error source, so \(\epsilon_{\Pi,\rm num}\) may
dominate the budget. Sliding the model along state \(\to\) lift \(\to\)
pixel while holding the \emph{certified object} fixed is the
representation-robustness axis of the paper.

\subsubsection{4.3 The certified object}\label{the-certified-object}

The object is the decoded physical invariant
\(\Psi(z) = H^\star(\Pi D_\psi z)\), and, for multi-invariant systems,
the decoded \emph{vector} invariant
\(\Psi_I(z) = \tilde I(\Pi D_\psi z)\) (for Kepler, the standardized
\((\tilde H,\tilde L)\)). We never certify the learned latent
Hamiltonian \(H_\theta\) or a learned scalar witness \(C_\omega\): both
are evaluated only as \emph{witnesses}, and a model can drive either to
be conserved while \(\Psi\) drifts. Figure 1 draws this explicitly ---
\(H_\theta,C_\omega\) are a side branch with no path to the certified
object, which is read off only where the state pipeline and the
latent-decoded pipeline meet on \(\Psi\). In practice \(\Psi\) is
evaluated step-by-step along the rollout: each latent
\(z_n = F_\theta^{\,n} E_\phi(o_0)\) is decoded and read out to
\((\hat q_n,\hat p_n) = \Pi D_\psi z_n\) and \(H^\star\) is evaluated
there, so the per-step decoded-energy drift
\(\delta_{\rm phys}^{\rm tube}\) and the initial defect \(\epsilon_0\)
are estimated as suprema over the calibration tube --- the
representation (\(\epsilon_0\)), readout (\(\epsilon_{\Pi,\rm num}\)),
and dynamics (\(\delta_{\rm phys}^{\rm tube}\)) terms of the budget.

\subsubsection{4.4 Model variants}\label{model-variants}

We compare three structural priors for the latent dynamics \(F_\theta\).
Within each representation level, variants use matched data, step size,
latent dimension, training budget, and evaluation pipeline (state has no
image encoder/decoder; lift and pixel do):

\begin{itemize}
\tightlist
\item
  \textbf{Plain.} An unconstrained residual latent map --- the
  no-structure baseline.
\item
  \textbf{Hard symplectic.} A latent integrator constrained to be
  symplectic (a separable Störmer--Verlet step and, separately, a
  SympNet), the standard inductive bias of Hamiltonian world models.
\item
  \textbf{Soft (B-self).} A self-supervised witness \(C_\omega\) ---
  scalar for energy shells, vector-valued for joint invariants
  (e.g.~\(C_\omega\in\mathbb{R}^2\) for Kepler) --- trained to be
  invariant along the latent rollout, with no access to \(H^\star\). It
  can support a certificate for the decoded invariant only through the
  alignment bridge of §4.6 --- the witness itself is never the certified
  object.
\end{itemize}

A \emph{privileged} controller that uses the true \((H,L)\) appears only
as a sanity reference / ablation, never as a method. Every reported
comparison fixes the certified object and varies only the prior and the
representation, so a horizon difference is attributable to structure,
not to a different evaluation target.

\subsubsection{4.5 The soft witness and its
training}\label{the-soft-witness-and-its-training}

The B-self witness is the load-bearing ``soft'' object, so we are
explicit about how it is trained --- and about what it is \emph{not}
given. Its objective combines (i) an \textbf{invariance-along-rollout}
term penalizing \(\lvert C_\omega(F_\theta z)-C_\omega(z)\rvert\); (ii)
a \textbf{true-temporal invariance} term tying \(C_\omega\) across real
trajectory observations,
\(\mathcal{L}_{\text{inv-data}} = \sum_k \lvert C_\omega(E_\phi o_{t+k}) - C_\omega(E_\phi o_t)\rvert^2\);
and (iii) a \textbf{variance / anti-collapse} term that prevents the
trivial constant solution. Crucially, \(H^\star\) never enters training:
the witness is discovered self-supervised and only \emph{afterwards}
aligned to decoded physical energy. The ablation in §5.3 confirms which
terms are load-bearing, and a battery of spurious controls --- shuffled,
random-scalar, trajectory-index, fresh-head --- collapse to negative
held-out \(R^2\), showing the alignment is real rather than a fitting
artifact. The witness is a small scalar (or, for joint invariants,
vector) head on the latent; its calibration and audit protocol are
specified in §4.6.

\subsubsection{4.6 Alignment-bridge
calibration}\label{alignment-bridge-calibration}

To turn the witness into a load-bearing bridge to the certified decoded
invariant we fit the calibration \(g_C\) as a
\textbf{controlled-Lipschitz monotone spline} on held-out energy shells,
and contrast it with an affine baseline and an uncontrolled isotonic
fit. The certified condition of Proposition C is bi-Lipschitz,
\(0<\kappa\le g_C'\le L_g\); empirically the spline is genuinely
bi-Lipschitz (finite \(L_g=0.606\), strictly positive \(\kappa=0.242\)),
whereas the isotonic fit has a clipped Lipschitz constant \(\sim 10^2\)
that drives \(T_{\rm align}\) to zero --- so \emph{monotone alone is
insufficient}. The bridge \(g_C\circ C_\omega\) is a witness alignment;
the certified object stays \(\Psi=H^\star(\Pi D_\psi z)\), and we report
the held-out audit \(R^2\) of \(g_C\circ C_\omega\) against \(\Psi\) as
the alignment quality. In the per-seed appendix tables this audit is
labeled \(\eta_C\) (after the witness \(C_\omega\)): the \(\eta_C\)
defect is the empirical counterpart of the decoder-side defect
\(\eta_D\) of §3.3 --- it is what enters \(T_{\rm align}\) and
\(\epsilon_\star = 2\eta_C/m\) --- and the ``\(\eta_C\) \(r^2\)'' column
reports the held-out audit \(R^2\).

\subsubsection{4.7 Tube-restricted certification and reporting
discipline}\label{tube-restricted-certification-and-reporting-discipline}

Every certificate is restricted to a tube \(K_{c,\rho}\) around the
shell, with the certifier
\(T_{\rm shell}=\lfloor(m_{c,\rho}\epsilon-\epsilon_0)/\delta_{\rm cert}\rfloor\)
matching Propositions A/B, and \textbf{when a sub-tube is used the
excluded fraction is reported}. For pixels we apply a frozen
\emph{readout-stable} rule: with a readout-Lipschitz score
\(s=L\cdot r\), calibration points with \(s\) above
\(\tau_s = 10\cdot\mathrm{median}_{\rm CAL}(s)\) are excluded, yielding
a readout-stable sub-tube on which the decoded invariant is faithful
(excluded fraction \(\le 2.3\%\) per seed). For the joint invariant a
rank guard (\(\operatorname{rank} D\tilde I = 2\), \(\sigma_*>0\))
prevents a degenerate direction from inflating the horizon and makes a
metric re-charting \emph{legal} only if it does not buy horizon by
collapsing \(\sigma_*\). Two disciplines hold throughout: every
certifier rule (the stable-tube rule, the \(\sigma_*\) guard, the
legal-chart anti-cheat; Appendix B) is \textbf{frozen before the final
run}, so a certificate cannot be tuned to pass; and pilot/diagnostic
quantities (e.g.~the four-way defect decomposition) show mechanism only
and are never quoted as final-table results. Horizons are read off a
frozen \(\epsilon\)-grid; a certificate is \emph{non-vacuous} when
\(T\ge 1\) for some grid \(\epsilon\), and we report per-seed
non-vacuous counts and the seed-level win rate \(P(T_B>T_{\rm plain})\)
rather than a single tuned point.

\begin{center}\rule{0.5\linewidth}{0.5pt}\end{center}

\subsection{5. Results}\label{results}

\subsubsection{5.1 The state / lift ladder (Figure
2)}\label{the-state-lift-ladder-figure-2}

\textbf{Figure 2} is the whole-paper results map: one representation per
column, each with its primary result, certificate status, and caveat.

\begin{figure}
\centering
\includegraphics[width=0.82\linewidth,height=\textheight,keepaspectratio,alt={State / lift / pixel certificate ladder: hard symplectic works in state (WS3a 5/5 vs 0/5); the soft invariant survives the learned lift while the hard prior does not; pixel certification is recovered on a readout-stable sub-tube.}]{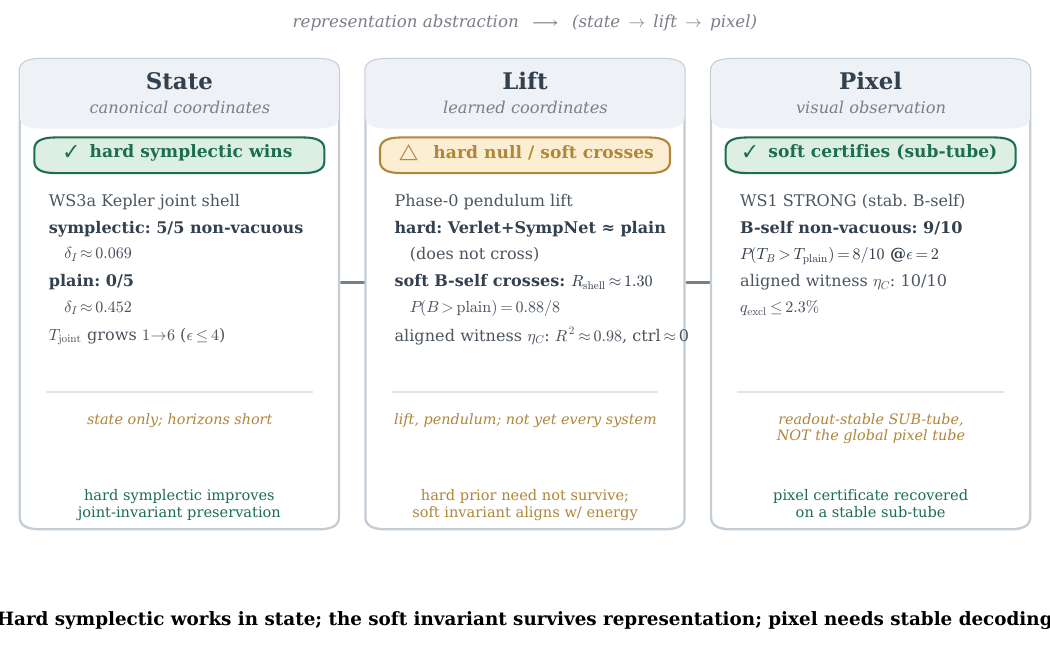}
\caption{State / lift / pixel certificate ladder: hard symplectic works
in state (WS3a \(5/5\) vs \(0/5\)); the soft invariant survives the
learned lift while the hard prior does not; pixel certification is
recovered on a readout-stable sub-tube.}
\end{figure}

\emph{State.} On state-space Kepler the joint-invariant shell behaves as
the theory predicts once train/eval eccentricity bands are aligned to
the certified tube: hard symplectic Verlet gives a robust, non-vacuous
\(T_{\rm joint}\) --- \textbf{symplectic \(5/5\) non-vacuous} (median
\(\delta_I = 0.069\)) versus \textbf{plain \(0/5\)}
(\(\delta_I = 0.452\)), a \(\approx 6.5\times\) smaller drift. A
privileged controller using the true \((H,L)\) is reported only as a
sanity reference. The horizons are short
(\(T_{\rm joint}(\epsilon{=}1) = 1\to 6\) over \(\epsilon\le 4\)): the
result is a \emph{robust structure separation}, not a long horizon.

\emph{Lift.} The move to a learned lift is where the hard-vs-soft story
appears. Hard symplectic structure does \textbf{not} cross the learned
chart --- a separable Verlet integrator and a SympNet are statistically
indistinguishable from plain (both at \(8\) seeds; SympNet/plain horizon
ratio \(1.03\)--\(1.04\) across the tolerance grid) --- while the soft
B-self invariant \textbf{does} cross, with a structure-gain ratio
\(R_{\rm shell}\approx 1.30\) and \(P(B>\mathrm{plain}) = 0.88\) over
\(8\) seeds, and a witness-alignment audit that is positive (mean
\(R^2 \approx 0.98\), controls \(\approx 0\)). The pattern generalizes:
on a second nonlinear, non-singular system (the Duffing double well,
same recipe, pre-registered gate) the soft invariant survives with a
\emph{larger} gain --- \(R_{\rm shell}\approx 2.3\),
\(P(B>\mathrm{plain})=0.88\) over \(8\) seeds, with both the per-step
drift and the representation defect improving --- while both hard arms
again fail to cross (ratios \(0.77\)--\(0.87\), \(P\le 0.5\)). On the
\emph{linear} oscillator lift, by contrast, all variants tie (ratios
\(\approx 1\)), so the soft gain appears with nonlinearity and grows
with it (oscillator \(\approx 1.0\times\), pendulum
\(\approx 1.3\times\), Duffing \(\approx 2.3\times\)). The
interpretation is Proposition D in action: symplecticity pins the form
of the latent flow but not the identity of the conserved scalar, so
under an arbitrary learned chart the symplectic prior no longer protects
\(H^\star\), whereas a witness trained to \emph{be} an invariant can be
aligned to it.

\subsubsection{5.2 The alignment bridge (Figure
3)}\label{the-alignment-bridge-figure-3}

\textbf{Figure 3} shows how a soft witness becomes a load-bearing bridge
to the certified decoded invariant, and what kind of calibration that
requires: \textbf{controlled-Lipschitz monotone}, not merely monotone.
On the pendulum, a monotone-spline calibration \(g_C\) beats an affine
baseline on held-out energy shells with no overfit: held-out
\(R^2 = 0.984\) (interleave) and \(0.988\) (extrapolate-outer) versus
affine \(0.950 / 0.967\). The calibration is genuinely bi-Lipschitz ---
finite \(L_g = 0.606\) and strictly positive \(\kappa = 0.242\) ---
exactly Proposition C's condition, and it yields a positive certified
horizon \(T_{\rm align} = 5.1\) steps at \(\epsilon = 1.0\), turning on
past a critical tolerance \(\epsilon_\star \approx 0.755\). Per-seed
\(T_{\rm align}\) is intrinsically high-variance (its denominator
\(L_g\,\delta_C\) can be small), so the load-bearing quantities are the
held-out \(R^2\) and the spline-versus-isotonic contrast, not the
absolute horizon.

\begin{figure}
\centering
\includegraphics[width=0.78\linewidth,height=\textheight,keepaspectratio,alt={WS2 alignment bridge: monotone alone is insufficient --- a controlled-Lipschitz (\textbackslash kappa=0.242, L\_g=0.606) spline calibration makes the decoded-invariant certificate non-vacuous, whereas uncontrolled isotonic monotonicity drives T\_\{\textbackslash rm align\}\textbackslash to0. The certified object remains H\^{}\textbackslash star(\textbackslash Pi D\_\textbackslash psi z).}]{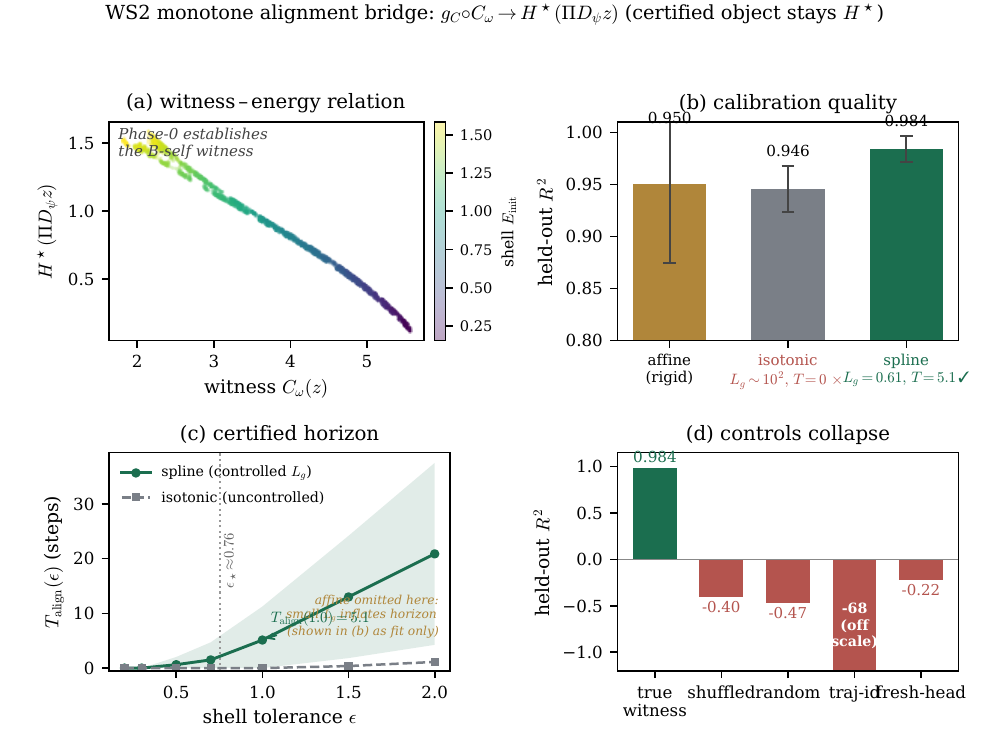}
\caption{WS2 alignment bridge: monotone alone is insufficient --- a
controlled-Lipschitz (\(\kappa=0.242\), \(L_g=0.606\)) spline
calibration makes the decoded-invariant certificate non-vacuous, whereas
uncontrolled isotonic monotonicity drives \(T_{\rm align}\to0\). The
certified object remains \(H^\star(\Pi D_\psi z)\).}
\end{figure}

Two controls make the claim load-bearing. The true witness beats every
spurious control by a wide margin: shuffled, random-scalar,
trajectory-index, and fresh-head calibrations all collapse to
\emph{negative} held-out \(R^2\) (\(-0.40\), \(-0.47\), \(-68\),
\(-0.22\)) --- alignment is real, not a fitting artifact. And the
spline-vs-isotonic contrast is the lesson: an isotonic monotone fit has
a clipped Lipschitz constant \(\sim 10^2\) and drives \(T_{\rm align}\)
to zero, so \emph{monotone alone is insufficient}. Throughout,
\(g_C\circ C_\omega\) is the bridge; the certified object stays
\(\Psi = H^\star(\Pi D_\psi z)\).

\subsubsection{5.3 Pixel decoded-energy recovery (Figure
4)}\label{pixel-decoded-energy-recovery-figure-4}

\begin{figure}
\centering
\includegraphics[width=0.92\linewidth,height=\textheight,keepaspectratio,alt={WS1 pixel decoded-energy recovery (full four-panel). (a) ladder attribution; (b) diagnostic \textbackslash delta-decomposition (soft-witness \textbackslash delta\_\{\textbackslash rm decoder\} 0.163 vs plain 2.16 below the frozen boundary \textbackslash approx0.22 --- a diagnostic, not a final metric); (c) the 10-seed certificate: 8/10 beat plain at \textbackslash epsilon=2.0, non-vacuous 9/10 vs plain 4/10, alignment-positive 10/10, \textbackslash delta\_\{\textbackslash rm tube\}\^{}\{\textbackslash rm stable\} 0.237 vs plain 0.371; (d) ablation: true-temporal invariance and anti-collapse load-bearing, stack decoder still recovers it. A certificate on the readout-stable sub-tube, not the global pixel tube.}]{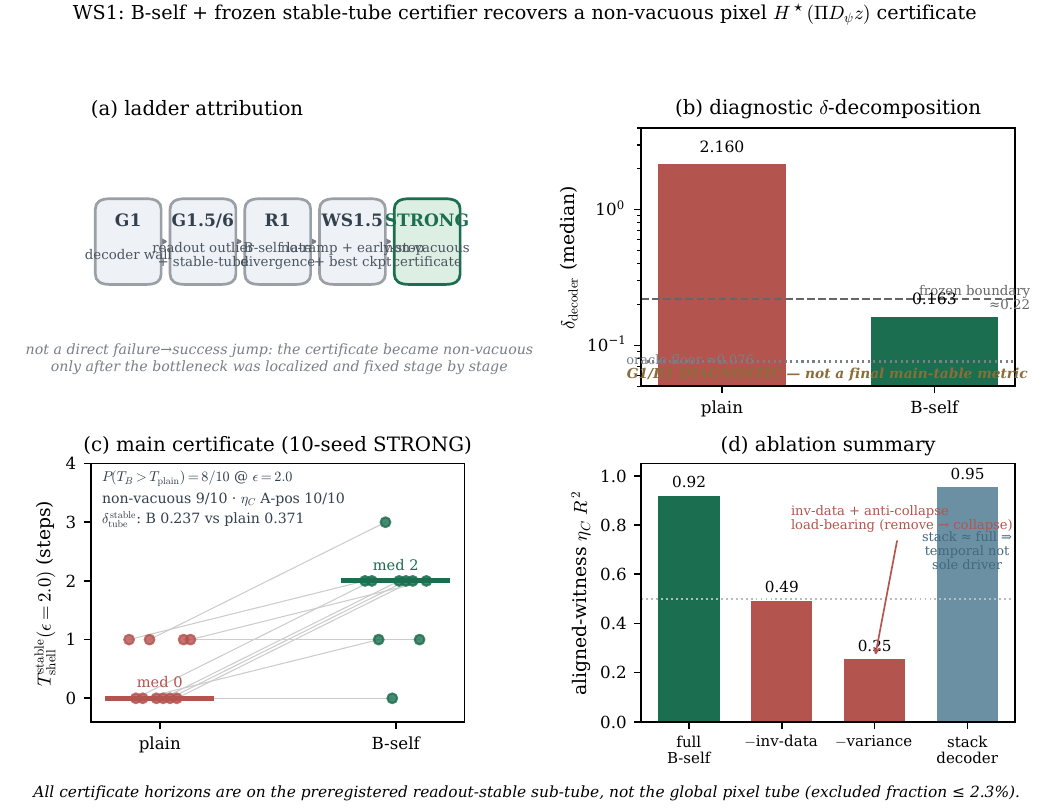}
\caption{WS1 pixel decoded-energy recovery (full four-panel). (a) ladder
attribution; (b) diagnostic \(\delta\)-decomposition (soft-witness
\(\delta_{\rm decoder}\) \(0.163\) vs plain \(2.16\) below the frozen
boundary \(\approx0.22\) --- a diagnostic, not a final metric); (c) the
10-seed certificate: \(8/10\) beat plain at \(\epsilon=2.0\),
non-vacuous \(9/10\) vs plain \(4/10\), alignment-positive \(10/10\),
\(\delta_{\rm tube}^{\rm stable}\) \(0.237\) vs plain \(0.371\); (d)
ablation: true-temporal invariance and anti-collapse load-bearing, stack
decoder still recovers it. A certificate on the readout-stable sub-tube,
not the global pixel tube.}
\end{figure}

Under pixel observation the \(\epsilon_{\Pi,\rm num}\) / decoder term
can dominate, and the question is whether the decoded-energy certificate
survives a \emph{visual} readout. On the readout-stable sub-tube, the
stabilized B-self witness recovers a non-vacuous pixel decoded-energy
certificate: \(P(T_B^{\rm stable}>T_{\rm plain}^{\rm stable}) = 8/10\)
at \(\epsilon = 2.0\) (and \(6/10\) at \(\epsilon=1.5\)), B-self
non-vacuous in \(9/10\) seeds versus plain \(4/10\), the alignment audit
positive in \(10/10\) (\(r^2\) \(0.88\)--\(0.96\)), with the tube defect
compressed to a median \(0.237\) versus plain \(0.371\), and a B-self
excluded fraction of at most \(2.3\%\) per seed. The honest scope is in
the name: this is a certificate on a \textbf{readout-stable sub-tube,
not the global pixel tube}. The ablation (Figure 4d) shows the drivers
are the soft-invariance losses and the certifier, not the decoder
architecture: removing true-temporal invariance drops the alignment
audit \(R^2\) to \(0.49\) and removing anti-collapse collapses it to
\(0.25\), whereas a stack decoder can recover the final certificate
under the stabilized recipe (\(R^2\approx0.95\)). So true-temporal
invariance and anti-collapse are load-bearing, while the temporal
decoder improves decoder consistency and was instrumental in diagnosing
the original pixel bottleneck rather than being the sole driver of
recovery.

\subsubsection{5.4 The Kepler geometric boundary (Appendix Figure
5)}\label{the-kepler-geometric-boundary-appendix-figure-5}

Kepler pushes the joint invariant \((\tilde H,\tilde L)\) through a
learned lift on the most nonlinear and only \emph{singular} system; the
result is a clean geometric boundary, established by a ladder that rules
out each easy explanation (full detail in Appendix Fig. 5 and Table
C.5). The state result already passes (§5.1: symplectic \(5/5\) vs plain
\(0/5\)). Under the learned lift the decoded-\(\tilde I\) reconstruction
defect exceeds the budget, and neither capacity nor geometry rescues it:
varying the latent width \(d_z\in\{8,16,32\}\) moves the floor only
\(1.21\times\), with the residual concentrated at small radius where
\(\partial H/\partial r\sim r^{-2}\) amplifies small \((q,p)\) errors;
and legal \((H,L)\) re-chartings cut the charted defect
\(\sim 2.8\times\) and even \emph{raise} \(\sigma_*\)
(\(0.097\to0.120\), \(\operatorname{cond}\le50\)) yet still give
\(T_J=0\). The conclusion is sharp and honestly attributed: the
Kepler-lift decoded joint-invariant certificate is floor-limited by
\textbf{near-periapsis \(\tilde H\)-stiffness}, robust to dynamics,
capacity, and legal metric geometry alike. \textbf{This is a
system-specific inverse-square geometric boundary; it is not evidence
that conservation-aligned certificates fail under learned lifts in
general, and it does not undercut the pendulum-lift / pixel positives of
§5.1--5.3.}

\begin{center}\rule{0.5\linewidth}{0.5pt}\end{center}

\subsection{6. Discussion and
limitations}\label{discussion-and-limitations}

We show that physical conservation can be certified through learned
representations only when the certified object is the decoded physical
invariant --- not a latent Hamiltonian or an unaligned learned witness.
The three settings tell one story about the \emph{representation
robustness} of certificates. In known canonical coordinates hard
symplectic structure is the clear winner: it buys the longest certified
shell because the integrator's conserved scalar \emph{is} --- up to
backward error --- the physical energy in those coordinates. Move to a
learned chart and that coincidence breaks, and Proposition D makes the
reason precise: symplecticity is a constraint on the \emph{form} of the
flow (it preserves a symplectic form), not a guarantee about
\emph{which} function is conserved. A symplectic learned map can
therefore conserve a shadow quantity that has drifted from \(H^\star\)
once the coordinates are learned --- exactly what the lift nulls show
(Verlet and SympNet collapse to plain), and the Kepler lift makes the
shadow concrete: there the symplectic variant conserves its own learned
latent \(H_\theta\) to a per-step drift of \(0.034\) while that scalar
is misaligned with the true \((H,L)\) through the representation. A
\emph{soft} invariant, by contrast, is trained to be an invariant of the
learned latents and can then be aligned to physical energy through the
bridge of §3.3; geometrically it behaves like a foliation of the latent
space by level sets that a monotone calibration can register, and so it
is more robust to the representation change that defeats the hard prior.

This is why the wording is careful: hard symplectic is \textbf{not}
useless. It is \emph{coordinate-sensitive} --- it wins where the chart
is right (the longest state-space horizons) and degrades where the chart
is learned, and it may remain essential for phase fidelity or other
geometric observables a scalar invariant does not capture. The
recommendation is therefore not ``drop symplectic structure'' but
``either certify in coordinates where the symplectic form is the
physical one, or pair the structure with an explicit alignment of the
conserved scalar.'' The pixel result adds an orthogonal axis: even with
a working soft witness, visual observation introduces a decoding/readout
bottleneck, and certification there is about stabilizing the readout
enough to find a sub-tube on which the decoded invariant is faithful ---
a perceptual problem, not a structural one. Kepler closes the arc on the
negative side: where the invariant itself is geometrically singular, no
amount of representation, capacity, or legal re-charting removes the
near-periapsis floor.

\subsubsection{6.1 Limitations and scope}\label{limitations-and-scope}

The positive certificates are demonstrated on low-dimensional
conservative systems and at modest horizons; the structure separations
are statistically robust but the certified horizons themselves are
short, and we make no claim about long-horizon or high-dimensional
rollouts. All certificates are \emph{tube-restricted}; the pixel
certificate is explicitly on a \textbf{readout-stable sub-tube} with a
small but nonzero excluded fraction (\(\le 2.3\%\)), not the global
tube. The soft-invariant survival result is shown for the lift on the
pendulum and the Duffing double well (with a linear-oscillator anchor on
which all variants tie), and for state-space Kepler; we do not claim it
for every system, and the Duffing run reports certificate-level gains
without a separate witness-establishment audit. The Kepler-lift result
is a \textbf{system-specific geometric boundary} driven by the
inverse-square periapsis singularity, reported with full ladder
attribution rather than as a structure failure. Finally, our
contribution is a certified-object \emph{methodology} and a
representation-robustness study, not a benchmark-scale vision world
model; we make no benchmark-leaderboard claim and ask to be read as the
former.

Future work is deliberately scoped \emph{out} of this paper. Engineering
an observation/lift representation in which near-periapsis \(\tilde H\)
is not stiff is a separate study; extending the joint-shell certificate
to a non-singular multi-invariant system would confirm the boundary is
geometric. The backward-error, action-drift, and shadowing results
(Propositions D--F) remain conditional; making the alignment bridge a
coordinate-free level-set stability theorem, and pinning down when a
symplectic learned map's shadow Hamiltonian aligns with
\(H^\star\circ\Pi\circ D\), are the open theory questions.

\begin{center}\rule{0.5\linewidth}{0.5pt}\end{center}

\subsection{References}\label{references}

\begin{center}\rule{0.5\linewidth}{0.5pt}\end{center}

\section{Appendix}\label{appendix}

\emph{Numbers are parsed from frozen tagged artifacts. Full anonymized
configuration dictionaries, run identifiers, and code are provided in
the supplementary material.}

\subsection{Appendix A. Proofs}\label{appendix-a.-proofs}

Throughout, \(\Psi(z)=H^\star(\Pi D_\psi z)\); \(K_{c,\rho}\) is a tube
around the shell \(\Sigma_c=\{x:H^\star(x)=c\}\); and
\(m_{c,\rho}=\inf_{x\in K_{c,\rho}}\lVert\nabla H^\star(x)\rVert>0\) is
the shell sensitivity (a regular-level assumption on the tube). A
generic distance lemma is used repeatedly: if
\(\lvert H^\star(x)-c\rvert\le\eta\) on the tube then
\(\mathrm{dist}(x,\Sigma_c)\le\eta/m_{c,\rho}\), since
\(\lVert\nabla H^\star\rVert\ge m_{c,\rho}\) (mean value theorem).
Horizons are integer step counts, hence the floor.

\textbf{A.1 Proposition A (state shell).} Let
\(\delta_{\rm state}=\sup_{x\in K}\lvert H^\star(f_\theta x)-H^\star(x)\rvert\)
and \(\epsilon_0^{\rm state}=\lvert H^\star(x_0)-c\rvert\). Telescoping
the one-step invariant change,
\[\lvert H^\star(x_n)-H^\star(x_0)\rvert \le \sum_{k=1}^{n}\lvert H^\star(f_\theta x_{k-1})-H^\star(x_{k-1})\rvert \le n\,\delta_{\rm state}.\]
By the distance lemma,
\(\mathrm{dist}(x_n,\Sigma_c)\le(\epsilon_0^{\rm state}+n\,\delta_{\rm state})/m_{c,\rho}\).
Requiring the bound \(\le\epsilon\) and solving for the largest integer
\(n\) gives
\(T_{\rm shell}^{\rm state}=\lfloor(m_{c,\rho}\epsilon-\epsilon_0^{\rm state})/\delta_{\rm state}\rfloor\).
\(\square\)

\textbf{A.2 Proposition B (latent decoded-energy shell).} Let
\(\hat x_n=\Pi D_\psi F_\theta^{\,n}E_\phi(o_0)\). Telescoping \(\Psi\)
along the latent rollout bounds the decoded-energy drift by
\(n\,\delta_{\rm phys}^{\rm tube}\), with
\(\delta_{\rm phys}^{\rm tube}=\sup_{z\in K_z}\lvert\Psi(F_\theta z)-\Psi(z)\rvert\).
The triangle inequality adds the initial embedding defect
\(\epsilon_0=\lvert H^\star(\Pi D_\psi E_\phi o_0)-H^\star(x_0)\rvert\)
and a readout/numerical term \(\epsilon_{\Pi,\rm num}\) (kept
\emph{separate} from the per-step decoder-consistency drift):
\[\lvert H^\star(\hat x_n)-H^\star(x_0)\rvert \le \epsilon_0+\epsilon_{\Pi,\rm num}+n\,\delta_{\rm phys}^{\rm tube}.\]
The distance lemma and solving for \(n\) give
\(T_{\rm shell}^{\rm latent}=\lfloor(m_{c,\rho}\epsilon-\epsilon_0-\epsilon_{\Pi,\rm num})/\delta_{\rm phys}^{\rm tube}\rfloor\).
The three additive terms are the \emph{representation} (\(\epsilon_0\)),
\emph{readout} (\(\epsilon_{\Pi,\rm num}\)), and \emph{dynamics}
(\(\delta_{\rm phys}^{\rm tube}\)) budget used throughout. \(\square\)

\textbf{A.3 Proposition C (alignment bridge).} Let \(C_\omega\) be a
(scalar or vector) witness with per-step drift
\(\delta_C=\sup\lVert C_\omega(Fz)-C_\omega(z)\rVert\), \(g_C\) an
\(L_g\)-Lipschitz calibration, and
\(\eta_D=\sup_z\lvert\Psi(z)-g_C(C_\omega z)\rvert\) the decoder-side
alignment defect. \emph{(C1, decoder-side.)} Inserting
\(g_C\circ C_\omega\) twice,
\[\lvert\Psi(F^nz)-\Psi(z)\rvert \le \underbrace{\lvert\Psi(F^nz)-g_C(C_\omega F^nz)\rvert}_{\le\,\eta_D} + \underbrace{\lvert g_C(C_\omega F^nz)-g_C(C_\omega z)\rvert}_{\le\,L_g\,n\,\delta_C} + \underbrace{\lvert g_C(C_\omega z)-\Psi(z)\rvert}_{\le\,\eta_D},\]
where the middle term uses Lipschitzness then telescoping. Hence
\(\lvert\Psi(F^nz)-\Psi(z)\rvert\le 2\eta_D+L_g\,n\,\delta_C\) and
\(T_{\rm align}=\lfloor(m_{c,\rho}\epsilon-2\eta_D)/(L_g\,\delta_C)\rfloor\).
\emph{(C2, physical-initial.)} Replacing one decoder-side \(\eta_D\) by
an encoder-side defect \(\eta_E\) (comparing
\(g_C(C_\omega E_\phi o_0)\) against \(H^\star(x_0)\)) gives the
physical-initial variant with \(\eta_D+\eta_E\) in place of \(2\eta_D\).
\emph{(Bi-Lipschitz role.)} The condition \(0<\kappa\le g_C'\le L_g\) is
load-bearing: \(L_g\) controls the bound while \(\kappa>0\) makes
\(g_C\) invertible, putting \(C_\omega\)'s level sets in bijection with
the decoded-energy level sets --- without it, a monotone-but-flat
calibration satisfies the inequality while certifying nothing (the
isotonic vacuity of §5.2). Throughout, \(g_C\circ C_\omega\) is a
\emph{bridge}; the certified object remains \(\Psi\). \(\square\)

\textbf{A.4 Joint-invariant shell.} For a vector invariant \(\tilde I\)
with per-step drift
\(\delta_I=\sup\lVert\tilde I(Fz)-\tilde I(z)\rVert\), telescoping gives
\(\lVert\tilde I(z_n)-\tilde I(z_0)\rVert\le\epsilon_0+n\,\delta_I\).
The guard \(\operatorname{rank}D\tilde I=2\) on the tube makes
\(\tilde I\) a submersion onto its joint level set, so
\(\sigma_*=\inf_{x\in K}\sigma_{\min}(D\tilde I(x))>0\) and the
joint-shell distance is controlled by
\(\lVert\text{drift}\rVert/\sigma_*\). Solving \(\le\epsilon\) gives
\(T_{\rm joint}=\lfloor(\sigma_*\epsilon-\epsilon_0)/\delta_I\rfloor\).
The guard makes a metric re-charting \emph{legal}: a chart cannot
inflate \(T_{\rm joint}\) by collapsing \(\sigma_*\), since
\(\sigma_*^J\) is recomputed through \(D\varphi\) (Appendix B.3).
\(\square\)

\textbf{A.5 Conditional statements (D--F).} Not load-bearing for the
empirical claims. \emph{(D, backward error / shadow.)} For a
near-identity analytic symplectic \(F_h\), a modified Hamiltonian
\(\widetilde H_h^{(N)}\) exists with
\(\phi^h_{\widetilde H}\approx F_h\) to \(O(h^{N+1})\); \emph{if}
\(\sup\lvert\widetilde H_h^{(N)}-H^\star\circ\Pi\circ D\rvert\le\eta_{\rm BEA}\),
a decoded-shell horizon follows (program-level). \emph{(E, action.)}
\(T_{\rm act}(\eta)\) targets the KAM/Nekhoroshev shape (a shape prior,
not a first-version theorem). \emph{(F, shadowing.)} Theorem-faithful
only on a uniformly hyperbolic toy; the double-pendulum case is
empirical-appendix only. Proposition D supplies the discussion's
intuition: \emph{symplecticity constrains the form of the flow, not
which scalar equals physical energy.}

\subsection{Appendix B. Frozen certifier
rules}\label{appendix-b.-frozen-certifier-rules}

Each rule is fixed \emph{before} the final run, so a certificate cannot
be tuned to pass.

\textbf{B.1 Pixel stable-tube certifier (v1.1).} With readout-Lipschitz
score \(s=L\cdot r\), exclude calibration points whose \(s_i\) exceeds
\(\tau_s=10\cdot\mathrm{median}_{\rm CAL}(s)\), giving the
readout-stable sub-tube. Frozen: the \(\tau_s\) rule, the
\(\epsilon\)-grid, the certifier
\(T_{\rm shell}=\lfloor(\epsilon m-\epsilon_0)/\delta_{\rm cert}\rfloor\)
(matching Prop A/B), the data splits, and the loss weights. Coverage:
the soft-witness excluded fraction is \(\le 2.3\%\) per seed (median
\(0.000\), max \(0.0234\)), re-pinnable from the per-seed certifier
JSON; the \emph{plain} variant's excluded fraction runs higher (up to
\(0.031\)), and the certificate uses the soft-witness coverage. (The
seed-0 certifier-definition figure \(0.008\) is a different quantity.)
The certificate is on this \textbf{readout-stable sub-tube, not the
global pixel tube.}

\textbf{B.2 Joint-shell rank guard.}
\(\sigma_*=\inf_{x\in K}\sigma_{\min}(D\tilde I)\) with
\(\operatorname{rank}D\tilde I=2\) on the tube; normalized
\(\tilde I=\big((H^\star-\mu_H)/s_H,\,(L-\mu_L)/s_L\big)\); the
\(\tilde I\)-calibration seed is frozen.

\textbf{B.3 Legal-chart anti-cheat.} A re-charting \(\varphi(H,L)\)
passes only if (frozen) \(T_J>0\) \textbf{and}
\(\sigma_*^J\ge0.3\,\sigma_*^{A0}\) \textbf{and}
\(\operatorname{cond}(D\varphi)\le50\). Guards: HL-only inputs;
model-independent; \(\sigma_*^J\) recomputed via \(D\varphi\); raw
baseline reported side-by-side; condition number capped.

\subsection{Appendix C. Seed-level
tables}\label{appendix-c.-seed-level-tables}

\textbf{C.1 State Kepler joint shell.} Pre-registered \(5\)-seed run
(identical e-band-aligned configuration as the original \(3\)-seed run
below; seeds \((0,1,2)\to(0\text{–}4)\), seeds \(0\)--\(2\) reproduce it
exactly). Main-text numbers use this run:

{\def\LTcaptype{none} 
\begin{longtable}[]{@{}
  >{\raggedright\arraybackslash}p{(\linewidth - 8\tabcolsep) * \real{0.2000}}
  >{\raggedright\arraybackslash}p{(\linewidth - 8\tabcolsep) * \real{0.2000}}
  >{\raggedright\arraybackslash}p{(\linewidth - 8\tabcolsep) * \real{0.2000}}
  >{\raggedright\arraybackslash}p{(\linewidth - 8\tabcolsep) * \real{0.2000}}
  >{\raggedright\arraybackslash}p{(\linewidth - 8\tabcolsep) * \real{0.2000}}@{}}
\toprule\noalign{}
\begin{minipage}[b]{\linewidth}\raggedright
variant
\end{minipage} & \begin{minipage}[b]{\linewidth}\raggedright
\(\delta_I\) median
\end{minipage} & \begin{minipage}[b]{\linewidth}\raggedright
\(\delta_I\) per-seed
\end{minipage} & \begin{minipage}[b]{\linewidth}\raggedright
non-vacuous
\end{minipage} & \begin{minipage}[b]{\linewidth}\raggedright
\(T_{\rm joint}\) at \(\epsilon=1.0/1.5/2.0/2.5/3.0/4.0\)
\end{minipage} \\
\midrule\noalign{}
\endhead
\bottomrule\noalign{}
\endlastfoot
plain & 0.452 & 0.388 / 0.489 / 0.732 / 0.452 / 0.424 & 0/5 &
0/0/0/0/0/1 \\
\textbf{symplectic} & \textbf{0.069} & 0.094 / 0.052 / 0.064 / 0.069 /
0.084 & 5/5 & 1/2/3/4/5/6 \\
conservation\_reg (privileged) & 0.075 & 0.127 / 0.061 / 0.082 / 0.075 /
0.075 & 4/5 & 1/2/3/3/4/6 \\
\end{longtable}
}

Original \(3\)-seed run (frozen tagged artifact; retained for
provenance):

{\def\LTcaptype{none} 
\begin{longtable}[]{@{}
  >{\raggedright\arraybackslash}p{(\linewidth - 8\tabcolsep) * \real{0.2000}}
  >{\raggedright\arraybackslash}p{(\linewidth - 8\tabcolsep) * \real{0.2000}}
  >{\raggedright\arraybackslash}p{(\linewidth - 8\tabcolsep) * \real{0.2000}}
  >{\raggedright\arraybackslash}p{(\linewidth - 8\tabcolsep) * \real{0.2000}}
  >{\raggedright\arraybackslash}p{(\linewidth - 8\tabcolsep) * \real{0.2000}}@{}}
\toprule\noalign{}
\begin{minipage}[b]{\linewidth}\raggedright
variant
\end{minipage} & \begin{minipage}[b]{\linewidth}\raggedright
\(\delta_I\) median
\end{minipage} & \begin{minipage}[b]{\linewidth}\raggedright
\(\delta_I\) per-seed
\end{minipage} & \begin{minipage}[b]{\linewidth}\raggedright
non-vacuous
\end{minipage} & \begin{minipage}[b]{\linewidth}\raggedright
\(T_{\rm joint}\) at \(\epsilon=1.0/1.5/2.0/2.5/3.0/4.0\)
\end{minipage} \\
\midrule\noalign{}
\endhead
\bottomrule\noalign{}
\endlastfoot
plain & 0.489 & 0.388 / 0.489 / 0.732 & 0/3 & 0/0/0/0/0/0 \\
\textbf{symplectic} & \textbf{0.064} & 0.094 / 0.052 / 0.064 & 3/3 &
1/2/3/4/5/7 \\
conservation\_reg (privileged) & 0.082 & 0.127 / 0.061 / 0.082 & 2/3 &
1/2/2/3/4/5 \\
\end{longtable}
}

Step-size hardening (\(h=0.02\to0.01\), same \(5\) seeds, pre-registered
option): the separation persists and sharpens --- symplectic \(5/5\)
non-vacuous (median \(\delta_I=0.032\),
\(T_{\rm joint}(\epsilon{=}1)=3\), curve \(3\to14\)) versus plain
\(0/5\) (\(\delta_I=0.184\)). The PASS is not a step-size artifact.

\textbf{C.2 Lift sweep} (pendulum).

{\def\LTcaptype{none} 
\begin{longtable}[]{@{}
  >{\raggedright\arraybackslash}p{(\linewidth - 10\tabcolsep) * \real{0.1667}}
  >{\raggedright\arraybackslash}p{(\linewidth - 10\tabcolsep) * \real{0.1667}}
  >{\raggedright\arraybackslash}p{(\linewidth - 10\tabcolsep) * \real{0.1667}}
  >{\raggedright\arraybackslash}p{(\linewidth - 10\tabcolsep) * \real{0.1667}}
  >{\raggedright\arraybackslash}p{(\linewidth - 10\tabcolsep) * \real{0.1667}}
  >{\raggedright\arraybackslash}p{(\linewidth - 10\tabcolsep) * \real{0.1667}}@{}}
\toprule\noalign{}
\begin{minipage}[b]{\linewidth}\raggedright
\(\epsilon\)
\end{minipage} & \begin{minipage}[b]{\linewidth}\raggedright
\(R_{\rm shell}^{B}\)
\end{minipage} & \begin{minipage}[b]{\linewidth}\raggedright
\(P(B{>}p)\)
\end{minipage} & \begin{minipage}[b]{\linewidth}\raggedright
\(R_{\rm shell}^{\rm sym}\)
\end{minipage} & \begin{minipage}[b]{\linewidth}\raggedright
\(P(s{>}p)\)
\end{minipage} & \begin{minipage}[b]{\linewidth}\raggedright
\(T_p\)/\(T_B\)/\(T_s\)
\end{minipage} \\
\midrule\noalign{}
\endhead
\bottomrule\noalign{}
\endlastfoot
0.30 & 1.41 & 0.88 & 1.03 & 0.38 & 4.5/6.2/4.6 \\
0.50 & 1.34 & 0.88 & 1.09 & 0.50 & 9.5/12.5/10.1 \\
0.70 & 1.32 & 0.88 & 1.09 & 0.50 & 14.5/19.0/15.2 \\
1.00 & 1.30 & 0.88 & 1.10 & 0.38 & 22.2/28.8/23.2 \\
\end{longtable}
}

Verdict: structural symplectic does \textbf{not} survive the lift
(sympl/plain \(\sim\!1.0\)); the soft witness \textbf{does}
(\(\sim\!1.3\times\) horizon, \(P{=}0.88\), 8 seeds). The
\textbf{witness-establishment audit} (10 seeds --- \emph{distinct} from
the WS2 calibration audit of C.3) gives mean \(R^2=0.982\) (per-seed
\(0.925\)--\(0.999\)), \(\lvert\rho\rvert\) \(0.995\)--\(0.999\), with
shuffled/random controls \(\approx0\); the WS2 fidelity gate reuses
seed-0 of this audit (\(R^2=0.99269\)). \textbf{SympNet arm (\(8\)-seed
confirmation).} A deterministic re-run of the same harness with a
fourth, non-separable symplectic arm (SympNet) reproduces the table
above exactly (identical seeds and configuration) and yields
SympNet/plain horizon ratios \(1.03/1.04/1.02/1.03\) at
\(\epsilon=0.30/0.50/0.70/1.00\): the non-separable hard prior collapses
to plain exactly as separable Verlet does, while the soft witness holds
\(1.29\)--\(1.39\).

\textbf{Second system (Duffing double well, pre-registered) and linear
anchor.} The same recipe on the Duffing double well (above-separatrix
band \(H\in[0.15,0.6]\); no critical point in the band; \(8\) seeds,
\(4\) arms) gives
\(T_{\rm plain}/T_B/T_{\rm sym}/T_{\rm sn} = 13.9/31.9/11.2/12.1\) at
\(\epsilon=1.0\): the soft gain is \emph{larger} than on the pendulum
(\(R_{\rm shell}^{B}=2.30\)--\(2.77\) across the grid,
\(P(B>\mathrm{plain})=0.88\) at every \(\epsilon\)) and two-sided ---
per-step drift \(4.9\to2.9\times10^{-2}\) \emph{and} representation
defect \(1.00\to0.55\times10^{-1}\) --- while both hard arms stay at or
below plain (\(0.77\)--\(0.87\), \(P\le0.5\)). On the \emph{linear}
oscillator lift (\(8\) seeds) all variants tie (\(0.96\)--\(1.05\); an
earlier single-seed \(R_{\rm shell}=2.2\) does not survive seeding), so
the soft gain appears with nonlinearity and grows with it: oscillator
\(\approx1.0\times\) → pendulum \(\approx1.3\times\) → Duffing
\(\approx2.3\times\). The Duffing run reports certificate-level
quantities only (no separate witness-establishment audit).

\textbf{C.3 WS2 alignment-bridge calibration} (8 seeds, interleave split
--- the \emph{calibration} audit, not C.2's witness-establishment
audit).

{\def\LTcaptype{none} 
\begin{longtable}[]{@{}
  >{\raggedright\arraybackslash}p{(\linewidth - 12\tabcolsep) * \real{0.1429}}
  >{\raggedright\arraybackslash}p{(\linewidth - 12\tabcolsep) * \real{0.1429}}
  >{\raggedright\arraybackslash}p{(\linewidth - 12\tabcolsep) * \real{0.1429}}
  >{\raggedright\arraybackslash}p{(\linewidth - 12\tabcolsep) * \real{0.1429}}
  >{\raggedright\arraybackslash}p{(\linewidth - 12\tabcolsep) * \real{0.1429}}
  >{\raggedright\arraybackslash}p{(\linewidth - 12\tabcolsep) * \real{0.1429}}
  >{\raggedright\arraybackslash}p{(\linewidth - 12\tabcolsep) * \real{0.1429}}@{}}
\toprule\noalign{}
\begin{minipage}[b]{\linewidth}\raggedright
seed
\end{minipage} & \begin{minipage}[b]{\linewidth}\raggedright
affine
\end{minipage} & \begin{minipage}[b]{\linewidth}\raggedright
isotonic
\end{minipage} & \begin{minipage}[b]{\linewidth}\raggedright
spline
\end{minipage} & \begin{minipage}[b]{\linewidth}\raggedright
\(L_g\)
\end{minipage} & \begin{minipage}[b]{\linewidth}\raggedright
\(\kappa\)
\end{minipage} & \begin{minipage}[b]{\linewidth}\raggedright
\(T_{\rm align}\) (spline)
\end{minipage} \\
\midrule\noalign{}
\endhead
\bottomrule\noalign{}
\endlastfoot
0 & 0.989 & 0.923 & 0.980 & 0.509 & 0.178 & 7 \\
1 & 0.995 & 0.984 & 0.991 & 0.473 & 0.177 & 7 \\
2 & 0.960 & 0.931 & 0.988 & 0.599 & 0.233 & 1 \\
3 & 0.780 & 0.924 & 0.961 & 0.917 & 0.125 & 0 \\
4 & 0.904 & 0.971 & 0.973 & 0.765 & 0.168 & 0 \\
5 & 0.994 & 0.924 & 0.986 & 0.566 & 0.366 & 1 \\
6 & 0.999 & 0.949 & 0.999 & 0.457 & 0.364 & 18 \\
7 & 0.984 & 0.959 & 0.997 & 0.562 & 0.321 & 7 \\
\textbf{mean} & \textbf{0.950} & \textbf{0.946} & \textbf{0.984} &
\textbf{0.606} & \textbf{0.242} & \textbf{5.1} \\
\end{longtable}
}

Extrapolate-outer: spline \(0.988\) / affine \(0.967\) / isotonic
\(0.944\). \(T_{\rm align}(\epsilon)\) sweep (interleave mean) over
\(\epsilon\in\{0.2,0.3,0.5,0.7,1.0,1.5,2.0\}\): spline
\(0/0/0.6/1.5/\mathbf{5.1}/13.0/20.9\); isotonic \(0/0/0/0/0/0.4/1.1\);
affine \(0/0/1.1/4.4/10.1/19.6/29.5\). \(\epsilon_\star\): spline
\(0.755\), isotonic \(1.188\), affine \(0.962\). Controls (held-out
\(R^2\)): shuffled \(-0.40\), random \(-0.47\), trajectory-index
\(-68\), fresh-head \(-0.22\). Fidelity \(0.99269298\) vs
\(0.99269301\). Affine's larger \(T_{\rm align}\) comes from a small
constant \(L_g\) on a worse fit; the certificate-grade contrast is
spline (controlled \(L_g\)) vs isotonic (uncontrolled, vacuous).

\textbf{C.4 Pixel recovery} (10 seeds; regenerated from per-seed JSONs).

{\def\LTcaptype{none} 
\begin{longtable}[]{@{}
  >{\raggedright\arraybackslash}p{(\linewidth - 14\tabcolsep) * \real{0.1250}}
  >{\raggedright\arraybackslash}p{(\linewidth - 14\tabcolsep) * \real{0.1250}}
  >{\raggedright\arraybackslash}p{(\linewidth - 14\tabcolsep) * \real{0.1250}}
  >{\raggedright\arraybackslash}p{(\linewidth - 14\tabcolsep) * \real{0.1250}}
  >{\raggedright\arraybackslash}p{(\linewidth - 14\tabcolsep) * \real{0.1250}}
  >{\raggedright\arraybackslash}p{(\linewidth - 14\tabcolsep) * \real{0.1250}}
  >{\raggedright\arraybackslash}p{(\linewidth - 14\tabcolsep) * \real{0.1250}}
  >{\raggedright\arraybackslash}p{(\linewidth - 14\tabcolsep) * \real{0.1250}}@{}}
\toprule\noalign{}
\begin{minipage}[b]{\linewidth}\raggedright
seed
\end{minipage} & \begin{minipage}[b]{\linewidth}\raggedright
\(T_B(1.5)\)
\end{minipage} & \begin{minipage}[b]{\linewidth}\raggedright
\(T_p(1.5)\)
\end{minipage} & \begin{minipage}[b]{\linewidth}\raggedright
\(T_B(2.0)\)
\end{minipage} & \begin{minipage}[b]{\linewidth}\raggedright
\(T_p(2.0)\)
\end{minipage} & \begin{minipage}[b]{\linewidth}\raggedright
B \(\delta_{\rm tube}^{\rm stable}\)
\end{minipage} & \begin{minipage}[b]{\linewidth}\raggedright
pl \(\delta_{\rm tube}^{\rm stable}\)
\end{minipage} & \begin{minipage}[b]{\linewidth}\raggedright
\(\eta_C\) \(r^2\)
\end{minipage} \\
\midrule\noalign{}
\endhead
\bottomrule\noalign{}
\endlastfoot
0 & 1 & 0 & 2 & 1 & 0.286 & 0.375 & 0.962 \\
1 & 1 & 0 & 2 & 0 & 0.226 & 0.286 & 0.884 \\
2 & 0 & 0 & 1 & 0 & 0.230 & 0.389 & 0.928 \\
3 & 2 & 1 & 3 & 1 & 0.185 & 0.366 & 0.938 \\
4 & 0 & 0 & 0 & 0 & 0.250 & 0.259 & 0.913 \\
5 & 1 & 0 & 2 & 0 & 0.305 & 0.381 & 0.905 \\
6 & 1 & 0 & 2 & 0 & 0.220 & 0.313 & 0.900 \\
7 & 1 & 0 & 2 & 0 & 0.231 & 0.586 & 0.893 \\
8 & 0 & 1 & 1 & 1 & 0.243 & 0.241 & 0.951 \\
9 & 1 & 1 & 2 & 1 & 0.280 & 0.422 & 0.918 \\
\end{longtable}
}

Aggregates: \(P(T_B>T_p)=8/10\) @ \(\epsilon{=}2.0\) (\(6/10\) @
\(1.5\); \(P(B\ge p)=10/10\)); B non-vacuous (\(\epsilon\ge1.5\))
\(9/10\) (plain \(4/10\)); \(\eta_C\) positive \(10/10\);
\(\delta_{\rm tube}^{\rm stable}\) B median \(0.237\) vs plain
\(0.371\). Diagnostic four-way decomposition (not a final metric): plain
\(\delta_{\rm decoder}\) \(2.16\) / soft-witness \(0.163\). Ablation
\(\eta_C\) \(R^2\) (3 seeds): full \(0.92\) / \(-\)true-temporal
\(0.49\) / \(-\)variance \(0.25\) / stack decoder \(0.95\).

\textbf{C.5 Kepler-lift rescue ladder.} Tier-2 (\(q,p\)-only weighting +
channel norm), per-seed \(\epsilon_{0,H}^{q95}\) \(4.63/5.90/4.88\)
(median \(4.88\); \((q,p)\) recon \(\sim0.066\)). Tier-2b capacity:
\(d_z\in\{8,16,32\}\to\epsilon_{0,H}^{q95}\) \(2.82/2.49/2.32\)
(\(1.21\times\)); periapsis bin (\(d_z{=}32\)) \(3.87\) vs \(1.22/1.17\)
at mid/large radius. Legal charts: baseline \(\epsilon_{0,J}\) \(11.69\)
/ \(\log(-H)\) \(5.63\) / \textbf{orbital \((a,\lambda)\) \(4.21\)}
(\(2.8\times\)), with \(\sigma_*\) \(0.097\to0.120\),
\(\operatorname{cond}\le50\), and all \(T_J=0\).

\subsection{Appendix D.
Reproducibility}\label{appendix-d.-reproducibility}

\begin{itemize}
\tightlist
\item
  \textbf{Artifacts.} The integration branch and per-phase release tags
  are listed in the supplementary material; 132 tests were green at the
  final merge.
\item
  \textbf{Seeds / splits.} Disjoint seed triples per run: train
  \texttt{seed}, eval \texttt{seed+99}, calibration \texttt{seed+777}.
  Seed counts: 8 (alignment bridge), 10 (pixel), 3 (Kepler
  state/lift/rescue). \texttt{use\_deterministic\_algorithms=True};
  BLAS/OMP threads pinned.
\item
  \textbf{Environments} (two per-run). (A) macOS-arm64: Python 3.11.15,
  PyTorch 2.12.0, NumPy 2.4.6, scikit-learn 1.9.0, SciPy 1.17.1. (B)
  Linux-x86\_64: Python 3.12.3, PyTorch 2.8.0+cu128, NumPy 2.1.2.
\item
  \textbf{Provenance.} Each run stores a config hash, commit, and full
  config in its per-seed JSON; the full anonymized configuration
  dictionaries are provided in the supplementary material. No filesystem
  paths, usernames, or branch names are embedded in the released
  artifacts.
\end{itemize}

\subsection{Appendix Figure 5 --- Kepler
boundary}\label{appendix-figure-5-kepler-boundary}

\begin{figure}
\centering
\includegraphics[width=0.95\linewidth,height=\textheight,keepaspectratio,alt={Kepler exposes the geometric price of decoded joint-invariant certification. (a) state joint shell (3-seed panel; the pre-registered 5-seed extension confirms 5/5 vs 0/5, Appendix C.1): symplectic 3/3 vs plain 0/3 (\textbackslash delta\_I 0.064 vs 0.489). (b) the autoencoder rungs as raw \textbackslash epsilon\_\{0,H\}\^{}\{q95\} and the legal-chart rung as the charted \textbackslash epsilon\_\{0,J\} --- the same quantity class on different chart metrics, separate budgets (not one axis). (c) the residual concentrates at small radius (periapsis 3.87 vs 1.22/1.17 at mid/large radius), exactly where \textbackslash partial H/\textbackslash partial r\textbackslash sim r\^{}\{-2\} amplifies small (q,p) errors. (d) legal (H,L) re-chartings improve the charted defect (2.8\textbackslash times) and even raise \textbackslash sigma\_*, yet all give T\_J=0. This is a system-specific inverse-square geometric boundary, not a structure failure.}]{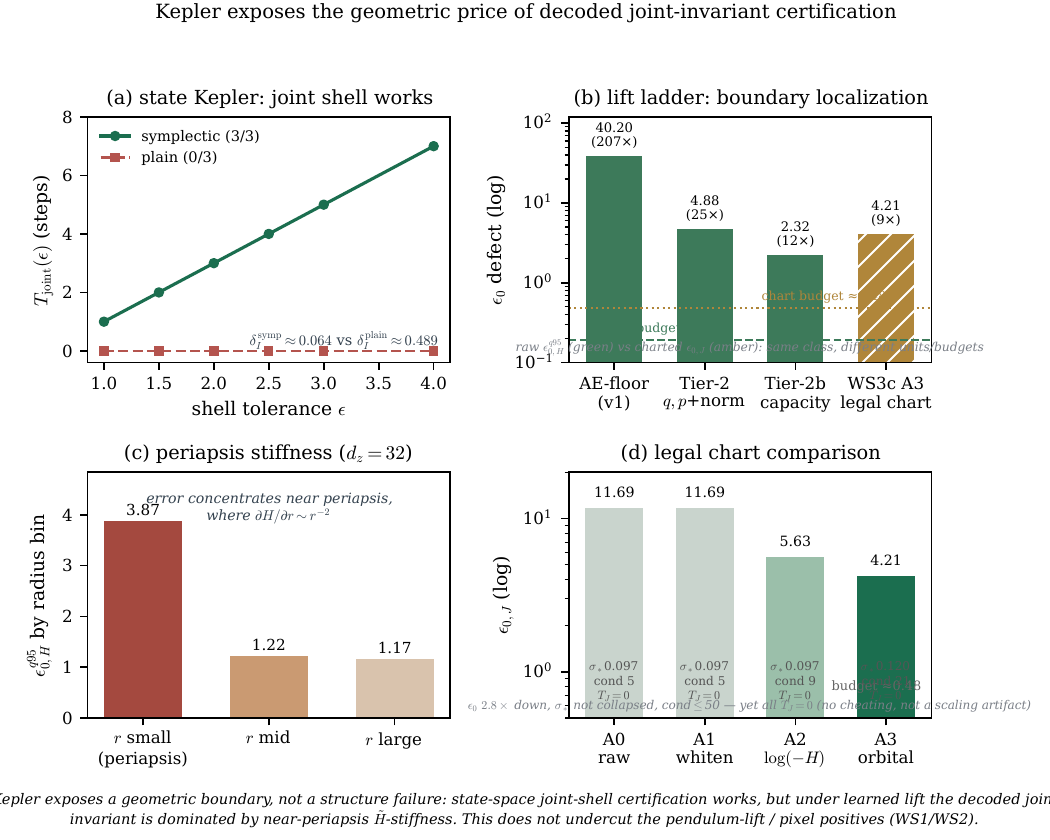}
\caption{Kepler exposes the geometric price of decoded joint-invariant
certification. (a) state joint shell (3-seed panel; the pre-registered
5-seed extension confirms \(5/5\) vs \(0/5\), Appendix C.1): symplectic
\(3/3\) vs plain \(0/3\) (\(\delta_I\) \(0.064\) vs \(0.489\)). (b) the
autoencoder rungs as raw \(\epsilon_{0,H}^{q95}\) and the legal-chart
rung as the charted \(\epsilon_{0,J}\) --- the same quantity class on
different chart metrics, separate budgets (not one axis). (c) the
residual concentrates at small radius (periapsis \(3.87\) vs
\(1.22/1.17\) at mid/large radius), exactly where
\(\partial H/\partial r\sim r^{-2}\) amplifies small \((q,p)\) errors.
(d) legal \((H,L)\) re-chartings improve the charted defect
(\(2.8\times\)) and even raise \(\sigma_*\), yet all give \(T_J=0\).
This is a system-specific inverse-square geometric boundary, not a
structure failure.}
\end{figure}

\begin{center}\rule{0.5\linewidth}{0.5pt}\end{center}

\bibliography{references}

\end{document}